\documentclass[sigconf]{aamas} 

\usepackage{fontawesome}
\usepackage{algorithm}
\usepackage[noend]{algorithmic}
\usepackage[switch]{lineno}

\DeclareMathAlphabet{\pazocal}{OMS}{zplm}{m}{n}

\usepackage{balance} % for balancing columns on the final page

\newtheorem{assumption}{Assumption}

\setcopyright{ifaamas}
\acmConference[AAMAS '26]{Proc.\@ of the 25th International Conference
on Autonomous Agents and Multiagent Systems (AAMAS 2026)}{May 25 -- 29, 2026}
{Paphos, Cyprus}{C.~Amato, L.~Dennis, V.~Mascardi, J.~Thangarajah (eds.)}
\copyrightyear{2026}
\acmYear{2026}
\acmDOI{}
\acmPrice{}
\acmISBN{}

\acmSubmissionID{1113}

\title[Sample-Efficient Neurosymbolic DRL]{Sample-Efficient Neurosymbolic Deep Reinforcement Learning}

\author{Celeste Veronese}
\affiliation{
  \institution{University of Verona}
  \city{Verona}
  \country{Italy}}
\email{celeste.veronese@univr.it}

\author{Alessandro Farinelli}
\affiliation{
  \institution{University of Verona}
  \city{Verona}
  \country{Italy}}
\email{alessandro.farinelli@univr.it}

\author{Daniele Meli}
\affiliation{
  \institution{University of Verona}
  \city{Verona}
  \country{Italy}}
\email{daniele.meli@univr.it}

\begin{abstract}
Reinforcement Learning (RL) is a well-established framework for sequential decision-making in complex environments. However, state-of-the-art Deep RL (DRL) algorithms typically require large training datasets and often struggle to generalize beyond small-scale training scenarios, even within standard benchmarks.
We propose a neuro-symbolic DRL approach that integrates background symbolic knowledge to improve sample efficiency and generalization to more challenging, unseen tasks. Partial policies defined for simple domain instances, where high performance is easily attained, are transferred as useful priors to accelerate learning in more complex settings and avoid tuning DRL parameters from scratch.
To do so, partial policies are represented as logical rules, and online reasoning is performed to guide the training process through two mechanisms: (i) biasing the action distribution during exploration, and (ii) rescaling Q-values during exploitation.
This neuro-symbolic integration enhances interpretability and trustworthiness while accelerating convergence, particularly in sparse-reward environments and tasks with long planning horizons.
We empirically validate our methodology on challenging variants of gridworld environments, both in the fully observable and partially observable setting. We show improved performance over a state-of-the-art reward machine baseline.
\end{abstract}


\newcommand{\BibTeX}{\rm B\kern-.05em{\sc i\kern-.025em b}\kern-.08em\TeX}

\begin{document}

%%% The following commands remove the headers in your paper. For final 
%%% papers, these will be inserted during the pagination process.

\pagestyle{fancy}
\fancyhead{}

%%% The next command prints the information defined in the preamble.

\maketitle 

%%%%%%%%%%%%%%%%%%%%%%%%%%%%%%%%%%%%%%%%%%%%%%%%%%%%%%%%%%%%%%%%%%%%%%%%

\section{Introduction}
Deep Reinforcement Learning (DRL) can be successfully applied to solve sequential decision-making problems, offering invaluable benefits for many real-world domains of application, e.g., robotic tasks \cite{ibarz2021train} and sustainability \cite{zuccotto2024reinforcement},  involving complex dynamics, multiple performance objectives, and large observation spaces.
However, DRL algorithms still present drawbacks that limit their wide adoption to real systems.
Firstly, DRL policies are black-box, which makes them hardly interpretable to humans, affecting trustworthiness and social acceptance \cite{vouros2022explainable}. 
Moreover, one big challenge of DRL lies in \emph{sample inefficiency} \cite{dulac2021challenges}, as the agent needs to collect numerous experiences from the environment to build an accurate model and achieve optimality.
Sample inefficiency is particularly problematic when scaling and generalizing DRL to environments with longer planning horizons, more sub-goals, and sparse rewards \cite{dulac2021challenges} (e.g., larger grids with more objects in gridworlds).
Though several approaches have been proposed to mitigate this issue, they either require the availability of a large variety of previous data \cite{bertran2020instance}, or rely on specific assumptions on the parametrization of the policy \cite{wang2019generalization}.
Heuristic-guided DRL can potentially mitigate the aforementioned issues, but the currently established approaches based on reward shaping or augmentation (reward machines \cite{icarte2018reward}) are practically sample-inefficient and sensitive to the accuracy of heuristics \cite{cheng2021heuristic}.

In this paper, we propose a novel neuro-symbolic approach to improve sampling efficiency of DRL when generalizing to domains with longer planning horizons, more sub-goals, and sparse rewards. Specifically, we adapt interpretable symbolic knowledge from small-scale, easy-to-solve scenarios to guide the training of DRL agents in more challenging settings, in which the neural algorithm obtains poor performance due to sample inefficiency.
We include symbolic knowledge as logical specifications (rules) representing an approximation of the policy learned by the agent in simpler domain instances (e.g., small grids with few objects in gridworlds), which require limited exploration. 
When facing more complex or diverse scenarios, the training of the DRL agent is then enhanced by performing autonomous reasoning on the knowledge transferred from the easier setting, in order to deduce the set of most promising actions given the observation of the agent.
We then leverage this knowledge as a planning heuristic to be used directly at the algorithmic level of DRL. In contrast to existing works requiring an exact definition of sub-goal specifications or sub-plans \cite{kokel2023reprel}, our methodology works even with imperfect symbolic knowledge (e.g., learned from data \cite{furelos2021induction}), and does not require DRL parameter re-tuning when generalizing large domain instances.
In more detail, the contributions of this paper are the following:
\begin{itemize}
    \item we propose a novel neuro-symbolic approach for DRL, which exploits autonomous reasoning on partial logical policy specifications (either handcrafted or acquired in easier settings) to deduce the most promising actions to be taken. We present two different integrations of the symbolic knowledge into the DRL training process, both in exploration and exploitation. Our solution is implemented for the popular class of $\epsilon$-greedy DRL algorithms, which are relevant for the DRL community \cite{liu2022understanding} but are still sample-inefficient \cite{dann2022guarantees}. %Moreover, $\epsilon$-greedy algorithms define a neat separation between exploration and exploitation, thus allowing an in-depth analysis of our neuro-symbolic implementation. 
    \item we exploit an $\epsilon$-decay strategy to balance between the neural and symbolic components without compromising the exploration-exploitation tradeoff of the original DRL algorithm;
    \item we assess the performance of our methodology in two relevant gridworlds, \emph{OfficeWorld} \cite{icarte2018reward} and \emph{DoorKey} \cite{MinigridMiniworld23}, characterized by sparse rewards, multiple sub-goals, and long planning horizons. In particular, we leverage logical heuristics learnt from DRL executions in small grids with few objects, and then assess sampling efficiency and training performance on larger grids with more items (hence, longer planning horizon and more sub-goals) and partially observable scenarios. We show that our methodology is more robust to imperfect partial policies, against an established neuro-symbolic DRL approach based on reward machines \cite{icarte2018reward}. We finally perform an ablation study to evidence the necessary balance between symbolic reasoning and DRL via both the $\epsilon$-decay mechanism and a \emph{confidence} parameter $\rho$ that assesses how much we trust the symbolic knowledge.
\end{itemize}

\section{Related Works}
The problem of sampling efficiency prevents DRL scalability and generalization in the presence of long planning horizons, sparse rewards, and many sub-goals.
Initially, ad-hoc algorithms have been developed to face this problem, such as Deep Q-Network (DQN), Rainbow DQN, SAC, etc. \cite{SuttonBarto2018}.
Some approaches then tackled this problem by increasing the variability of the environment at the training stage, possibly with past training information \cite{bertran2020instance} or human intervention \cite{strouse2021collaborating}. 
%This solution, however, requires significant computational resources and does not exploit knowledge derived from past training in small domains to generalize to larger settings.
In \cite{igl2019generalization,lee2020network} regularization techniques are employed to prevent overfitting in DRL, which is a major cause for the lack of generalization. Nevertheless, these approaches still require numerous training iterations in large domains; furthermore, the interpretability of the learned policies is still hindered, especially when they are represented by deep neural networks.

The methodology proposed in this paper lies in the neuro-symbolic DRL research area, which combines the abstraction and generalization capabilities of interpretable symbolic (logical) representation and reasoning tools with the inherent advantages of neural approaches when dealing with uncertain data from environmental interaction.
The most prominent approach to neuro-symbolic DRL exploits the definition of logical specifications to derive automata, driving the agent towards sub-goals in a hierarchical planning framework \cite{kokel2023reprel} or by shaping the reward \cite{de2019foundations, furelos2021induction}. 
%However, these approaches require significant and highly reliable prior task knowledge, which may be unavailable in complex scenarios. 
However, reward shaping approaches still require many environmental interactions and do not solve the sample-inefficiency, especially with a long planning horizon or with imperfect heuristics \cite{cheng2021heuristic}. 
Recent works \cite{meli2024learning,sreedharan2023optimistic,veronese2024} propose to learn symbolic knowledge from past traces (state-action pairs) of an agent's executions, instead of handcrafting symbolic knowledge. 
However, they focused either on the exploration phase only \cite{meli2024learning, veronese2024}, or on a soft initialization of Q-values in tabular RL \cite{sreedharan2023optimistic}, without comprehensively realizing an efficient and adaptive neuro-symbolic integration for DRL.
% Moreover, \cite{meli2024learning} focused on model-based RL, while \cite{veronese2024, sreedharan2023optimistic} on tabular RL, thus limiting the applicability of the proposed solutions. 
% Despite these limitations, learning logical specifications remains the most promising direction, since it requires much less prior domain knowledge, only relying on the definition of a set of relevant task concepts, which does not necessarily describe the entire environment \cite{meli2024learning}.
On the contrary, we propose a novel neuro-symbolic methodology for DRL, which jointly leverages the symbolic knowledge at the algorithmic level, both in exploitation and exploration. Specifically, we consider the popular class of $\epsilon$-greedy DRL algorithms, where exploitation and exploration are neatly separated. We modulate the exploration factor $\epsilon$ to probabilistically favour the selection of symbolically entailed actions. This is achieved by biasing the action distribution in exploration and adaptively re-scaling the Q-values during exploitation, according to $\epsilon$-parameter. Thanks to an $\epsilon$-decay strategy, the early exploration of the agent is more sample-efficient, a fundamental requirement for scalable DRL \cite{jiang2024importance}. 
At the same time, we progressively modulate the impact of symbolic reasoning over DRL, mimicking the human-like synergy between fast and slow thinking \cite{kautz2022third}.
% and preserve the asymptotic optimality guarantees, even in the presence of imperfect symbolic knowledge.

Finally, our methodology presents some similarities with Statistical Relational Learning (SRL) \cite{marra2024statistical}, as proposed in \cite{hazra2023deep}. However, SRL hardly scales to complex domains, requiring the accurate definition of the policy search space \cite{hazra2023deep}. Furthermore, by exploiting the advantages of both symbolic reasoning and DRL, with respect to pure logical learning proposed in \cite{hazra2023deep}, our algorithm efficiently generalizes to more challenging domains with longer planning horizons and more sub-goals, where standard DRL algorithms fail.

\section{Background}
We here introduce the relevant background to our methodology, i.e., solving Markov Decision Processes (MDPs) with $\epsilon$-greedy DRL, our testing domains, and the logical framework of Answer Set Programming (ASP) \cite{lifschitz1999answer}, which is the state of the art for logical representation and reasoning on planning problems \cite{meli2023logic}.

\subsection{MDPs and Reinforcement Learning}
Markov Decision Processes (MDPs) provide a formal framework for planning and decision-making in deterministic and stochastic environments\cite{SuttonBarto2018}. A MDP is defined by the tuple \((S, A, T, R, \gamma)\), where \(S\) is the set of states describing the environment; \(A\) is the set of possible actions; \(T : S \times A \rightarrow \Pi(S)\) is the state transition function, representing the probability of transitioning to state \(s'\) from state \(s\) when action \(a\) is taken; \(R : S \times A \rightarrow \mathbb{R}\) is the reward function, providing the immediate reward received after taking action \(a\) in state \(s\); \(\gamma \in [0, 1]\) is the discount factor, which determines the present value of future rewards.
Planning with an MDP aims at finding an optimal policy \(\pi^*: S \to A\) that maximizes the expected cumulative reward (return) over time. The return, starting from state \(s\) and following a policy \(\pi\), is captured by the value function \(V^\pi(s)\):
\[
V^\pi(s) = \mathbb{E}\left[\sum_{t=0}^{\infty} \gamma^t R(s_t, a_t) \,|\, s_0 = s, \pi \right]
\]
% The optimal value function \(V^*(s)\), which gives the maximum expected reward from each state, satisfies the Bellman optimality equation:
% \[
% V^*(s) = \max_{a \in A} \left[R(s, a) + \gamma \sum_{s' \in S} T(s'|s, a)V^*(s') \right]
% \]
% By solving this equation, the optimal policy \(\pi^*\) can be derived, guiding the agent to take actions that maximize long-term rewards.

In model-free DRL, the agent learns to solve a MDP by interacting with the environment directly, learning an optimal policy and the value function as it performs actions in the environment and collects rewards.
Typically, DRL is based on the interleave between exploitation and exploration. During \emph{exploitation}, the agent selects the action according to the currently learned policy and value models; in the \emph{exploration} phase, the agent randomizes its decision in order to explore new regions of the policy space and ultimately improve its learned model.
In this paper, we focus on a specific class of DRL algorithms based on $\epsilon$-greedy exploration strategy \cite{liu2022understanding}, where typically a random action is chosen with $\epsilon$ probability.
DQN is the most popular representative of this class of algorithms \cite{SuttonBarto2018}. 
It uses a deep neural network (Q-network) to approximate an action-value function \( Q(s, a; \theta) \), where \(s\) represents the state, \(a\) the action, and \(\theta\) the parameters of the neural network.
% The Q-network estimates the utility to perform an action at a given state.
% Parameters $\theta$ are updated to minimize the difference between the predicted action value and a target value computed from the immediate reward \(r_{t+1}\) and the estimated maximum future reward at the next state \(s_{t+1}\), using a separate target network with parameters \(\theta^*\). The loss function is defined as:
% \[
% L(\theta) = \mathbb{E}\left[\left( r_{t+1} + \gamma \max_{a'} Q(s_{t+1}, a'; \theta^*) - Q(s_t, a_t; \theta) \right)^2\right]
% \]
% where \(\gamma\) is the discount factor that controls the trade-off between immediate and future rewards.
DQN employs an \(\epsilon\)-greedy policy with an \(\epsilon\)-decay strategy to balance exploration and exploitation, by gradually reducing $\epsilon$ to favor exploitation as the training progresses.
Initially, \(\epsilon\) is set to a high value, encouraging exploration by selecting random actions. Over time, \(\epsilon\) is gradually reduced according to a decay schedule, allowing the agent to exploit the learned Q-values more frequently as training progresses. This decay is crucial for the agent to sufficiently explore the environment in the early stages and focus on exploiting its knowledge later. 
% While $\epsilon$-greedy algorithms have demonstrated success in solving complex problems with large action and state spaces \cite{gimelfarb2020epsilon,du2021bilinear,liu2022understanding}, defining an effective and efficient exploration strategy is still a major issue \cite{dann2022guarantees}. This is a crucial limitation in MDP domains with sparse rewards, requiring a long planning horizon and sub-goal achievement under partial observability, where efficient exploration is needed to learn a good generalizable policy.

\subsection{Domains}\label{sub:domains}
\begin{figure}
    \centering
{\includegraphics[width=0.5\linewidth]{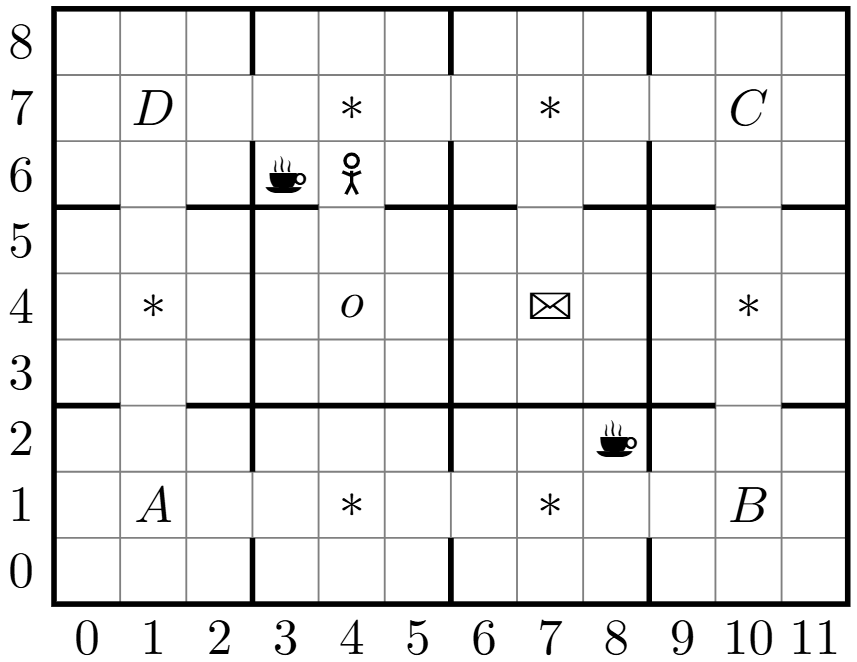}} \hfill
{\includegraphics[width=0.4\linewidth]{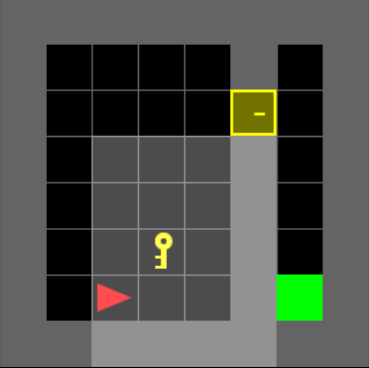}}
    \caption{Our testing domains: \emph{OfficeWorld} (left) and \emph{DoorKey} (right).}\label{fig:domains}
\end{figure}

In the \textbf{OfficeWorld} domain (Figure \ref{fig:domains}) \cite{icarte2018reward}, the agent is moving on a $9\times12$ grid, representing various rooms. The state space is given by the set of all locations in the grid (i.e., the agent knows the coordinate of the location it is stepping on), and the observable elements represented by coffee, mail, decorations (marked as $*$ in Figure \ref{fig:domains}), rooms $A,B,C,D$ and office $o$.
The fully deterministic action space is defined as $A = \{ \mathtt{left}, \mathtt{right}, \mathtt{up}, \mathtt{down}\}$.
We tested our methodology on the \emph{DeliverCoffeeAndMail} task, which requires bringing both coffee and mail to the office, and on the \emph{PatrolABC} task, in which the agent must visit rooms A, B, and C to complete the task. In both cases, stepping on a decoration ends the episode with a failure.

In the \textbf{DoorKey} domain (Figure \ref{fig:domains}), an agent (red) has to pick up a key to unlock a door (the key and the door must be of the same color) and then get to the green goal square. 
The environment is discretized in uniform cells, and the state space the agent can observe is a $7\times 7$ portion of the full grid in front of it (unless walls are present), as shown in light gray in Figure \ref{fig:domains}. Each tile is encoded as a 3-dimensional tuple \texttt{(object, color, state)}, indicating, respectively, the object that occupies that cell (e.g. a key, a door or a wall), its color and its state (open or closed, in case the object is a door). 
The action space for the domain is $A = \{\mathtt{left, right, forward, pickup, open}\}$. Performing actions $\mathtt{left}$ and $\mathtt{right}$, the agent turns in the corresponding direction; the $\mathtt{forward}$ action makes the agent move one cell ahead in the current direction; $\mathtt{pickup}$ is only effective in front of a key and it results in picking the key up; and $\mathtt{open}$ can be used by the agent to open a door when in front of it, by using a previously collected key. 
Both environments are fully deterministic, and a reward of $1-0.9 * (\mathtt{step\_count} / \mathtt{max\_steps})$ is given for completing the task, $0$ otherwise.

Both domains are particularly challenging because of the highly sparse reward and the need to coordinate multiple macroactions towards sub-goals (e.g., picking the key before opening the door and reaching the goal) for the winning strategy. 

\subsection{Answer Set Programming and Reasoning}
Answer Set Programming (ASP) \cite{lifschitz1999answer} represents a planning domain by a sorted signature $\pazocal{D}$, with a hierarchy of symbols defining the alphabet of the domain (variables, constants, and predicates). 
Logical axioms or rules are then built on $\pazocal{D}$.
In this paper, we leverage ASP formalism to represent the background knowledge about the MDP policy, either defined by human experts or learned via DRL in small scenarios. Hence, we consider normal rules, i.e., axioms in the form \( \mathtt{h :- b_1, \dots, b_n} \), where the body of the rule \( \mathcal{B} = \mathtt{b_1} \land \dots \land \mathtt{b_n} \) (i.e. the logical conjunction of the literals) serves as the precondition for the head \( \mathtt{h} \). 
In our setting, body literals represent features generated from the state of the environment (e.g., $\mathtt{samecolor(X,Y)}$ denoting that two elements are of the same color in \emph{DoorKey}), while the head literal represents an action (e.g., $\mathtt{pickup}$ in \emph{DoorKey}).

Given an ASP problem formulation $P$, an ASP solver computes the \emph{answer sets}, i.e., the minimal models satisfying ASP axioms, after all variables have been \emph{grounded} (i.e., assigned with constant values). Answer sets lie in Herbrand base \( \mathcal{H}(P) \), defining the set of all possible ground terms that can be formed. In our setting, answer sets contain the feasible actions available to the DRL agent.

\section{Methodology}
Our neuro-symbolic DRL strategy (SR-DQN) combines $\epsilon$-greedy DRL exploration and exploitation with Symbolic Reasoning (SR) over logical knowledge, approximating a good partial policy for small and simple domain instances.
For simplicity, we frame our methodology in the context of the well-established DQN algorithm; however, it can be easily extended to any other $\epsilon$-greedy DRL approach (e.g. Dueling DQN \cite{wang2016dueling}, Rainbow DQN \cite{hessel2018rainbow}.
We now detail the main phases of SR-DQN.

\subsection{Logical Representation of the MDP}
We begin by representing the MDP domain using the logical formalism of ASP. This involves encoding the state and action spaces as ASP terms. In the ASP signature, we define terms corresponding to environmental features $\pazocal{F}$, which capture essential aspects of the state (e.g., $\mathtt{locked(X)}$ in the \emph{DoorKey} domain, indicating that door $\mathtt{X}$ is locked), as well as terms for actions $\pazocal{A}$ (e.g., $\mathtt{left}$, a constant term representing the corresponding action).

To bridge the MDP and the corresponding logical representations, we introduce a \emph{feature map} $F_\pazocal{F} : S \rightarrow \pazocal{H(F)}$ and an \emph{action map} $F_\pazocal{A} : A \rightarrow \pazocal{H(A)}$, where $\pazocal{H(F)}$ and $\pazocal{H(A)}$ denote the Herbrandt bases of $\pazocal{F}$ and $\pazocal{A}$, respectively. 
More specifically, the feature map $F_\pazocal{F}$ translates an MDP state $s$ into a logical description composed of ground terms from $\pazocal{F}$. For example, in the example environment depicted in Figure \ref{fig:domains}, a specific state $s$ might be represented as $F_{\pazocal{F}}(s) = {\mathtt{key(X), notcarrying}}$. Likewise, the action map $F_\pazocal{A}$ provides a logical representation of MDP actions using ground terms from $\pazocal{A}$.
This part of our methodology relies on the following assumptions:

\begin{assumption}\label{ass:feature} The sets $\pazocal{A}$, $\pazocal{F}$ and the mappings $F_\pazocal{F}$ and $F_\pazocal{A}$ are known a priori. \end{assumption}

This is a standard assumption in the literature \cite{meli2024learning,furelos2021induction}, and it is considerably weaker than requiring full symbolic task specifications or the existence of a symbolic planner \cite{kokel2023reprel}. In particular, $F_\pazocal{A}$ can be seen as a straightforward symbolic encoding of the MDP's action space. Furthermore, we do not assume that $\pazocal{F}$ is \emph{complete}, i.e., it need not contain all task-relevant predicates. This relaxation is justified because our neuro-symbolic integration is robust to imperfect partial policies (see Section \ref{sec:met_nesy}). Thus, we assume that a minimal set of domain predicates, along with a grounding mechanism $F_{\pazocal{F}}$, is either known or can be obtained via automated symbol grounding techniques \cite{umili2023grounding}.

\begin{assumption}\label{ass:invert} The action map $F_\pazocal{A}$ is surjective. \end{assumption}

In other words, to each ground logical predicate corresponds at least one MDP action.
This assumption is typically satisfied in practice. For discrete action spaces, distinct predicates can be assigned to each action (i.e., $F_\pazocal{A}$ is bijective). In continuous settings, it is possible to define a discretization of the action space, where each grounded action predicate maps to a set of continuous actions. It is also possible to learn this mapping \cite{umili2023grounding}.

\subsection{Logical Representation of Policy Knowledge}\label{sec:log_repr}
Once the MDP is expressed in ASP formalism, we can represent the background information about the policy in terms of the maps $F_{\pazocal{F}, A}$. To this aim, we define a \emph{partial logical policy} $\pi_{ASP} : \pazocal{F} \rightarrow \pazocal{A}$, which maps environmental features to action terms. The logical policy encodes normal rules in the form $\mathtt{a :- f_1, \ldots, f_n}$, with $f_i \in \pazocal{F}, a \in \pazocal{A}$.
For instance, in \emph{DoorKey} domain, $\pi_{ASP}$ may correspond to the rule:
\begin{equation*}
    \mathtt{pickup(X) :- key(X), door(Y), samecolor(X,Y).}
\end{equation*}
meaning that the agent should pick up a key if it is of the same colour as the door.
% These rules can either be defined by a human with partial domain knowledge \cite{sreedharan2023optimistic}, but more importantly, they can be learned from previous experience (state-action pairs) collected by the agent in small domain settings \cite{hazra2023deep,meli2025inductive}.

We remark that this knowledge can be inaccurate, e.g., learned from previous example executions \cite{furelos2021induction}.
In order to account for the possible inaccuracy of partial policies, we also define a confidence level $\rho \in [0,1)$ about $\pi_{ASP}$.
% Given a state $s$ of the MDP, we can then translate it into the corresponding set of ASP features through $F_\pazocal{F}$, and perform Answer Set reasoning (e.g. as proposed in \cite{gebser2019multishot}) to obtain the answer set that satisfies the set of axioms $\pazocal{K}$, representing the encoded knowledge. We can then employ $F^{-1}_{\pazocal{A}}$ to convert the actions included in the answer set back into MDP actions, obtaining suggestions regarding the best actions to perform according to the previous experience in the domain.

\subsection{Neuro-symbolic training}\label{sec:met_nesy}

\begin{algorithm}[ht]
\caption{SR-DQN training loop}
\label{alg:sr_dqn}
\begin{algorithmic}[1]
\REQUIRE Env, maps $F_\pazocal{F}$, $F_\pazocal{A}$, partial policy $\pi_{ASP}$, 
max episodes $E$, max steps $T$, exploration parameters $\epsilon_i$, $\epsilon_f$, $\epsilon_r$, partial policy confidence $\rho \in [0,1)$

    \STATE Initialize replay memory $M$, $Q(s,a;\theta)$, $\epsilon \gets \epsilon_i$
    \FOR{episode $e = 1$ to $E$} 
        \STATE Initialize state $s_0$
        \FOR{step $t = 1$ to $T$}
            \STATE Uniform sample $x \sim [0,1]$
            \IF{$x \geq \epsilon$}
                \STATE $a \gets$ \textcolor{red}{SR-Exploitation($s_t$, $\epsilon$, $\rho$)}
            \ELSE
                \STATE $a \gets$ \textcolor{red}{SR-Exploration($s_t$, $\epsilon$, $\rho$)}
            \ENDIF
            \STATE $s_{t+1}, r \gets$ Env.step(a)
            \STATE Store $(s_t, a, r, s_{t+1})$ in $M$
            \STATE $\theta \gets$ \textbf{DQNUpdate}($\theta$)
            \STATE $s_t \leftarrow s_{t+1}$
            \IF{$s_t$ is terminal}
                \STATE \textbf{break}
            \ENDIF
        \ENDFOR
        \IF{$\epsilon > \epsilon_f$}
        \STATE \textcolor{red}{$\epsilon \gets$ LinearDecrease($\epsilon, \epsilon_f, \epsilon_r$)}
        \ENDIF
    \ENDFOR
\end{algorithmic}
\end{algorithm}

\noindent Algorithm \ref{alg:sr_dqn} gives a general view of how we integrate symbolic reasoning into DQN. The main components of our methodology are highlighted in red.
Our goal is to improve the outcome of $\epsilon$-greedy DRL, by reasoning over the logical policy $\pi_{ASP}$ in order to efficiently bias the agent towards the most promising actions from background policy knowledge.
As in standard DQN, after initializing the Q-network $Q$ and the replay buffer $M$, we set the exploration rate $\epsilon$ to a suitable high initial value $\epsilon_i$ (Line 1), in order to favour exploration at the early stages of training, where the agent has not collected enough experience to build an accurate Q-network. During each episode of training, starting at state $s_0$ (Line 3), at each time step $t$, the algorithm samples a random number $x \in [0,1]$ uniformly (Line 5), to decide whether to perform exploitation (Line 6-7) or exploration (Line 8-9), both of which are improved via symbolic reasoning and will be explained in detail in the next subsections. 
After an action is taken, the agent observes the next state and reward, storing this experience in the replay memory for later training. The Q-network is then updated according to the original DRL algorithm. Finally, once the episode reaches termination (Line 15), the exploration rate $\epsilon$ is decreased until the lower bound $\epsilon_f$ is reached (Line 17)\footnote{The linear decrease rule is chosen empirically in our methodology, but more sophisticated strategies can be adopted to reduce $\epsilon$, depending on the specific task \cite{gimelfarb2020epsilon}.}. 

The exploration fraction parameter $\epsilon_r$ controls how quickly, during the training, $\epsilon_f$ must be reached, in terms of the number of episodes. For example, $\epsilon_r = 0.5$ means that $\epsilon = \epsilon_f$ is reached at episode $e = E / 2$.
In this way, the agent favours exploitation in place of exploration as the training progresses, while also relying more on the knowledge gained by the network and less on the background logical knowledge derived from smaller domains.
%Depending on the task, the update of $\epsilon$ can be performed at \emph{batches} of episodes, rather than after every single one. The linear decrease rule is chosen empirically in our methodology, but more sophisticated strategies can be adopted to reduce $\epsilon$, depending on the specific task \cite{gimelfarb2020epsilon}.

We now explain in detail how we employ symbolic reasoning in the different phases of training and then analyze how these changes affect the optimality guarantees of the original DRL algorithm.

\subsubsection{Neuro-symbolic exploration}\label{sec:nesy_exploration} 
\begin{algorithm}[ht]
\caption{SR-Exploration}
\label{alg:sr_exp}
\begin{algorithmic}[1]
\REQUIRE $F_\pazocal{F}$, $F_\pazocal{A}$, $\pi_{ASP}$,current state $s_t$, $\epsilon$, $\rho$
    \STATE $\pazocal{A}_{\pi_{ASP}} \gets$ \textbf{ComputeAnswerSet($\pi_{ASP}, F_\pazocal{F}(s_t)$)} 
    \STATE $A_{\pi_{ASP}} \gets F^{-1}_{\pazocal{A}}(\pazocal{A}_{\pi_{ASP}})$
    \IF{$A_{\pi_{ASP}} \neq \emptyset$}
        \STATE sample $a \sim$ \textbf{WeightProb($A, A_{\pi_{ASP}}, \rho$)} 
    \ELSE
        \STATE Uniform sample $a \sim A$
    \ENDIF
    \RETURN $a$
\end{algorithmic}
\end{algorithm}
 
\noindent The exploration phase in standard $\epsilon$-greedy approaches consists of picking a uniformly random action from the set $A$. On the contrary, in SR-Exploration (as shown in Algorithm \ref{alg:sr_exp}) automated reasoning is performed over $\pi_{ASP}$ to identify the set $\pazocal{A}_{\pi_{ASP}}$ of actions (ground terms in ASP formalism) entailed by the background policy knowledge, given the current set of ground environmental features $F_{\pazocal{F}}(s_t)$ (Line 1). Suggested ASP ground actions are then translated to the MDP action space $A_{\pi_{ASP}} \subseteq A$, by considering the pre-image $F^{-1}_{\pazocal{A}}$ of the action map (Line 2, see Assumption \ref{ass:feature}). The agent then selects an action from $A$ according to a weighted probability distribution (Line 4), where the weights are defined for each action as follows:
\begin{equation}
    w_a =
    \begin{cases}
        \rho \quad &\text{if } a \in A_{\pi_{ASP}}\\
        1-\rho \quad &\text{otherwise}
    \end{cases}\label{eq:weights}
\end{equation}
and then normalized, such that $\sum_{a \in A} w_a = 1$.
As explained in Section \ref{sec:log_repr}, $\rho$ represents the level of confidence about the knowledge encoded in $\pi_{ASP}$, which may be inaccurate, especially when it is learned \cite{meli2024learning}. 
The higher $\rho$ value, the higher the probability that the agent selects an action suggested by background policy knowledge.
If $A_{\pi_{ASP}} = \emptyset$ (i.e., $\pi_{ASP}$ cannot suggest any valuable action at the given state $s_t$), then a uniformly random action is selected from $A$ as in standard DQN (Line 6).

\subsubsection{Neuro-symbolic exploitation}
Our approach, shown in Algorithm \ref{alg:sr_qvals}, aims at enhancing the standard DRL exploitation phase by biasing the choice towards the most promising action according to the symbolic knowledge.
Given the current state $s_t$, the agent first queries the Q-network $Q$ to obtain estimated Q-values for all possible actions (Line 1). As in DQN, these Q-values represent the expected return for each action under the current policy. We then employ the ASP policy $\pi_{\text{ASP}}$ in the same way as explained in Section \ref{sec:nesy_exploration} to compute the set of preferred actions $\pazocal{A}_{\pi_{\text{ASP}}}$ (Line 2).
Subsequently, the Q-values are rescaled (Line 3) by a factor $k_a = 1+(\epsilon*w_a)$ for each action, with $w_a$ determined following Equation \ref{eq:weights}.
This is aimed at adjusting the action values within the context of the most promising action set $\pazocal{A}_{\pi_{\text{ASP}}}$, according to the confidence parameter $\rho$. Adding $\epsilon$ as an additional rescaling parameter allows the agent to increasingly trust the estimations produced by the neural network as the training proceeds. 
Finally, the action with the highest rescaled Q-value is selected for execution (Line 4).

\begin{algorithm}
\caption{SR-Exploitation}
\label{alg:sr_qvals}

\begin{algorithmic}[1]
\REQUIRE $F_\pazocal{F}$, $F_\pazocal{A}$, $\pi_{ASP}$, access to current network $Q$,
current state $s_t$, $\epsilon$, $\rho$ 
    \STATE Qvals $\gets Q(s_t, a; \theta)$
    \STATE $\pazocal{A}_{\pi_{ASP}} \gets$ \textbf{ComputeAnswerSet($\pi_{ASP}, F_\pazocal{F}(s_t)$)} 
    \STATE $\text{Qvals}_{\pi_{ASP}}\gets$ \textbf{RescaleQvals}(Qvals, $\pazocal{A}_{\pi_{ASP}}, \rho$)
    \STATE $a \gets \arg\max_{a} \text{Qvals}_{\pi_{ASP}}$
    \RETURN $a$
\end{algorithmic}
\end{algorithm}

\subsection{Impact of symbolic knowledge on training convergence}
The integration of symbolic knowledge we propose is designed to improve the efficiency of DRL exploration while maintaining the stability and empirical convergence behavior of the base algorithms. This balance is achieved by modulating the influence of symbolic guidance through the exploration factor ($\epsilon_r$, $\epsilon_f)$, which gradually decreases as training progresses, and the confidence parameter $\rho$.
At the beginning of training, when the agent has limited experience and the Q-value estimates are still inaccurate, the symbolic component plays a stronger role in shaping the agent's behavior. By biasing the action selection toward more promising actions, the agent can more efficiently explore relevant regions of the state space, thereby accelerating early learning.
As $\epsilon_t$ decays, the contribution of symbolic guidance becomes progressively weaker, allowing the learned value function to increasingly dominate the action selection process. This gradual reduction prevents the logical policy from over-constraining the agent’s behavior or trapping it in suboptimal local minima that may arise from incomplete or imperfect symbolic knowledge.
The symbolic component thus acts as a form of guided exploration in the early stages, while the long-term learning dynamics of the underlying DRL algorithm remain unaffected. 

\subsection{Complexity of symbolic inference in training}
At each training step (Lines 7 an 9 of Algorithm \ref{alg:sr_dqn}), the RL agent needs to evaluate whether $F_\pazocal{F}(s)$ (i.e. the grounding of the current MDP state) satisfies $\pi_{ASP}$, which is a fixed set of non-disjunctive ASP rules.
It is known that the grounding part of ASP reasoning is the most computationally intensive \cite{gebser2019multishot}.
Specifically, let each normal rule $r \in \pi_{ASP}$ include $v$ variables that range over a finite domain of $N$ constants. Denote the finite set of all possible ground instances as $g(r, F_\pazocal{F}(s))$. The number of such ground instances is $O(N^v)$.
Verifying whether these instances are satisfied by $F_\pazocal{F}(s)$ then requires checking, for each ground instance, the truth of every literal in its body. For propositional (ground) non-disjunctive programs, this process is linear in the number of literal occurrences \cite{eiter1997abduction}. Therefore, the total computational cost of verifying all grounded instances of rules in $\pi_{ASP}$ can be expressed as:
\[
T_{\text{ASP}}(F_{\pazocal{F}}(s),\pi_{ASP}) = O\!\left(\sum_{r \in \pi_{ASP}} |g(r,F_{\pazocal{F}}(s))| \cdot |\mathcal{B}_r|\right),
\]
where $|\mathcal{B}_r|$ denotes the number of literals composing the body of rule $r$.
In our domain representations, each rule references only a small subset of state predicates (e.g., objects within the agent’s view), thus the combinatorial term $|g(r, F_\pazocal{F}(s))|$ remains small in practice. 
In more general and complex settings, we remark that $\pi_{ASP}$ doesn't evolve during training and $\pazocal{F}$ is known a priori. Hence, it is possible to compute the full grounding before training starts, thereby eliminating the exponential factor of the complexity.
Consequently, the symbolic reasoning over $\pi_{ASP}$ introduces an overhead which is proportional to the number of variable instantiations actually realized in the current state. For compact rule sets and moderate grounding sizes, this practically results in a negligible increment of the original DQN step time.

\section{Empirical Evaluation} \label{sec:exp}
We evaluate our SR-DQN methodology on the \emph{DoorKey} and \emph{OfficeWorld} domains presented in Section \ref{sub:domains}.
These domains are widely used in related literature \cite{hazra2023deep,icarte2018reward,furelos2021induction}, since they present unique challenges, namely i) sparse reward definition; ii) long planning horizon; iii) the need for optimal strategic coordination towards the achievement of sub-goals. Hence, they represent the ideal benchmark to validate our methodology.

In the following, we primarily evaluate the scalability and sampling efficiency of our method in both domains, increasing the planning horizon and number of sub-goals. To this aim, we compare with a standard DQN baseline and a state-of-the-art reward machine methodology as proposed by \cite{icarte2018reward}.
We then perform a thorough ablation study in the \emph{DoorKey} domain (which was the most challenging one in our experiments), to separately investigate the performance of symbolic exploration vs. exploitation, and to assess the impact of relevant parameters of our methodology. 
Importantly, we tuned the DQN baseline on the easier settings (e.g., maps with one key only in Doorkey) and applied it to more challenging scenarios using the same set of hyperparameters. This allowed us to test the capabilities of our approach in achieving generalization without requiring the DRL algorithm to be tuned from scratch.

\subsection{Logical Domain Representations and Policies}
We now introduce the symbolic knowledge $\pi_{ASP}$ for our testing domains. Crucially, symbolic knowledge represents partial policies which are learned by the authors of \cite{furelos2021induction, hazra2023deep} in small-scale domains. Hence, they may fail to generalize to more complex settings, e.g., with more items and sub-goals and a longer planning horizon.

\subsubsection{OfficeWorld}
We formalize the \emph{OfficeWorld} domain by introducing environmental features $\pazocal{F}$ that represent the known positions of the observables in the map:  $\mathtt{coffee(X)}$, $\mathtt{mail(Y)}$, $\mathtt{office(Z)}$. Moreover, we introduce the $\mathtt{hasCoffee}$ and $\mathtt{hasMail}$ predicates to state that the agent has already picked up that item, and \\$\mathtt{hittingDecoration}$, which represents the presence of a decoration (that must not be broken) in the agent's moving direction. Finally, predicate $\mathtt{visited(X)}$ states that the agent has already visited room $\mathtt{X}$.
As logical policies, we take the ones learned by \cite{furelos2021induction}. Namely, the policy for the \emph{DeliverCoffee} task:
\begin{align}
    \mathtt{goto(X) :-} &\mathtt{\,\, coffee(X),\,\, not\,\, hasCoffee,} \label{eq:coffee}\\\nonumber&\mathtt{not\,\, hittingDecoration.}\\
    \mathtt{goto(X) :-} &\mathtt{office(X),\,\, hasCoffee,}\label{eq:office}
    \\\nonumber&\mathtt{not\,\, hittingDecoration.}
\end{align}
and the one for the \emph{VisitAB} task:
\begin{align}
    \mathtt{goto(A) :-} &\mathtt{\,\, \,\,visited(NONE).} \label{eq:visitA}\\
    \mathtt{goto(B) :-} &\mathtt{\,\,visited(A).}\label{eq:visitB}
\end{align}
where $\pazocal{A} = \{\mathtt{goto(X)}\}$ denotes the action of moving to an item. For Assumption \ref{ass:invert}, we map this to 
\begin{align*}
    &\mathtt{left :-} \mathtt{\,\, goto(X), on\_left(X).} \\
    &\mathtt{right :-} \mathtt{\,\, goto(X), on\_right(X).} \\
    &\mathtt{forward :-} \mathtt{\,\, goto(X), straight(X).}
\end{align*}
adding the necessary body predicates to $\pazocal{F}$.
The above specifications suggest that the agent should first pick up the coffee (Rule (\ref{eq:coffee})) and then reach the office (Rule (\ref{eq:office})), both without breaking any decoration.

\subsubsection{Doorkey}
For simplicity, we replicate the ASP formulation of the Doorkey domain proposed in \cite{hazra2023deep}. As environmental features $\pazocal{F}$, we define $\mathtt{door(X)}, \mathtt{key(Y)}, \mathtt{goal(Z)}$ to denote doors, keys and the goal. We then introduce predicate $\mathtt{locked(X)}$ to state that a door is locked; $\mathtt{unlocked}$ to state the intermediate door to the goal is open; $\mathtt{samecolor(X,Y)}$ to denote items (either doors or keys) of the same colour; $\mathtt{carrying(Y)}$ to specify that the agent is carrying an object (key); and $\mathtt{notcarrying}$ to specify that the agent is not carrying anything. 

We also consider the logical policy learned in \cite{hazra2023deep}, in a small $5\times 5$ grid with only one door and key:
\begin{align}
    \mathtt{pickup(X) :-} &\mathtt{\,\, key(X), samecolor(X,Y),} \label{eq:pickup}\\\nonumber&\mathtt{door(Y), notcarrying}\\
    \mathtt{open(X) :-} &\mathtt{key(Z), samecolor(X,Z),}\label{eq:open}
    \\\nonumber&\mathtt{door(X), locked(X), carrying(Z)} \\
    \mathtt{goto(X) :-} &\mathtt{\,\, goal(X), unlocked} \label{eq:goto}
\end{align}
Rule \eqref{eq:pickup} suggests that the agent should pick up a key $\mathtt{X}$ if it matches the colour of the door $\mathtt{Y}$ and the agent is not currently carrying another key. Then, from Rule \eqref{eq:open}, the agent can unlock a door $\mathtt{X}$ if it is holding a matching-colored key $\mathtt{Z}$, the door $\mathtt{X}$ is locked and  $\mathtt{Z}$ is the correct key for that door.
Finally, Rule \eqref{eq:goto} prescribes that the agent moves towards the goal 
$\mathtt{X}$ if all doors along the path are unlocked.
% Differently from \cite{hazra2023deep}, we consider a partially observable environment.

\subsection{Scalability study}
We compare SR-DQN against the performance of a standard DQN algorithm (DQN in the figures) and DQN with reward machines (RM-DQN in the figures) as designed in \cite{icarte2018reward}. 
For reward machines, we test different rewards for state transitions in both Doorkey and OfficeWorld and keep the best-performing ones in the tuning scenarios.
For SR-DQN, we empirically choose $\epsilon_f=\epsilon_r=0.3$ in Doorkey tasks, and $\epsilon_f=0.05$, $\epsilon_r=0.1$ in OfficeWorld tasks.
Since we do not have information about the confidence level of $\pi_{ASP}$ from \cite{hazra2023deep} and \cite{furelos2021induction}, we empirically set $\rho=0.8$ for both domains.
For each method, we evaluate the discounted return\footnote{For RM-DQN, we exclude the additional reward from the plots for a fair comparison.} achieved over 5 random seeds.

\begin{figure*}[ht]
    \centering
    \includegraphics[width=0.32\linewidth]{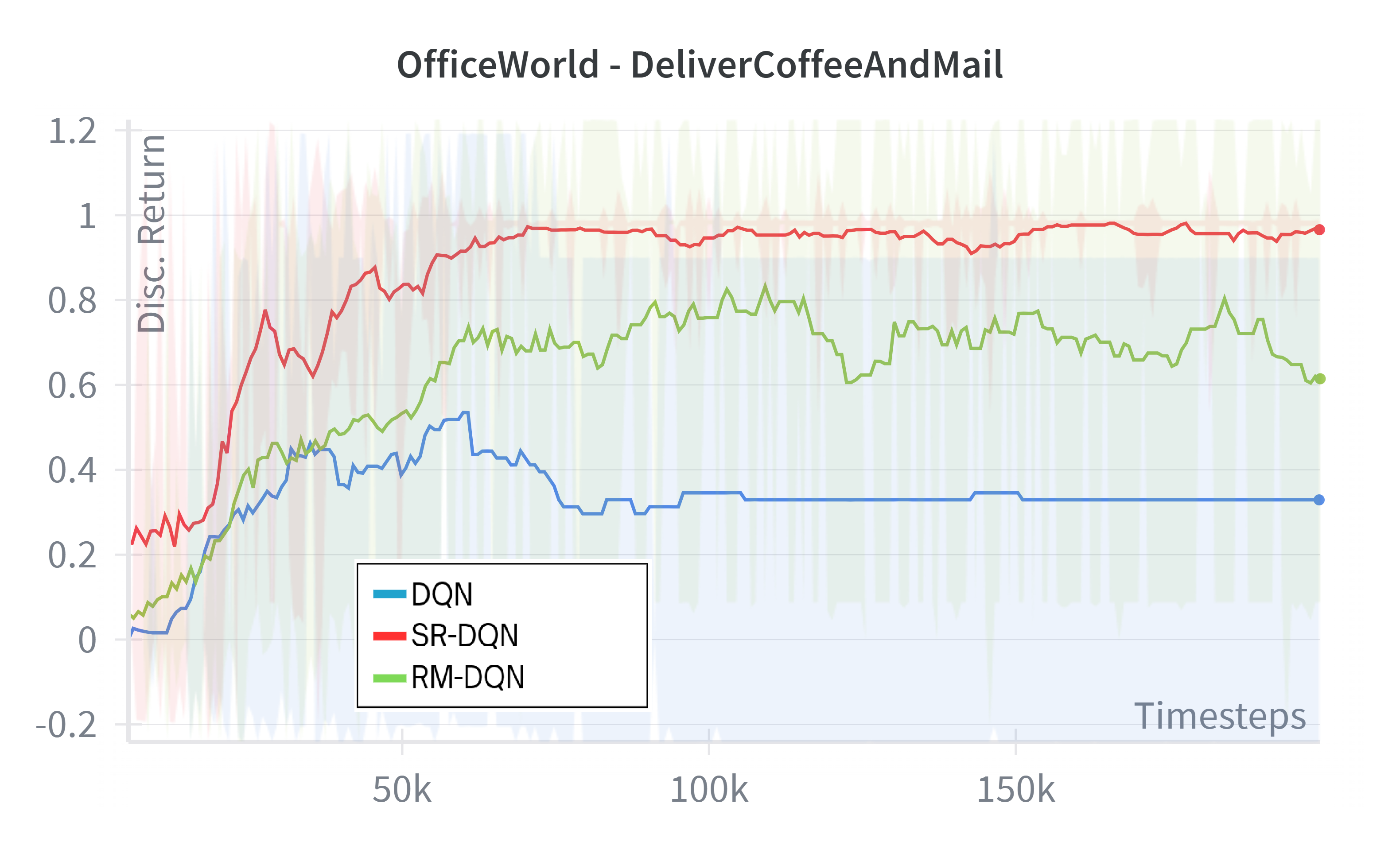}
    \label{fig:office_tasks}
    \hspace{1cm}
    \includegraphics[width=0.32\linewidth]{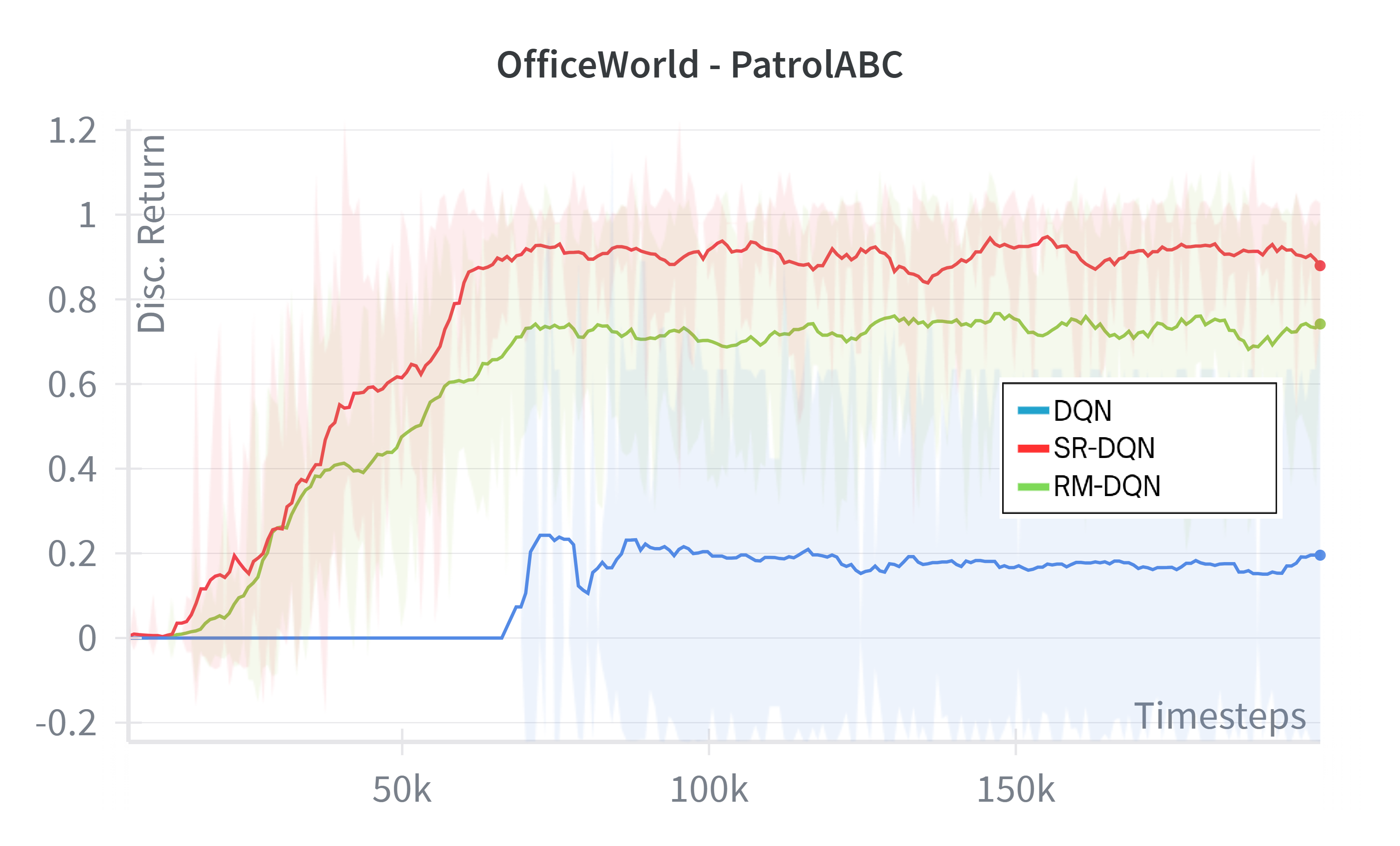}
    \caption{\label{fig:office}OfficeWorld results on the \emph{DeliverCoffeeAndMail} task (left) and on the \emph{PatrolABC} task (right).}
\end{figure*}

\subsubsection{OfficeWorld}
To test the scalability performance of SR-DQN in the \emph{OfficeWorld} domain, we employ the policy learned by \cite{furelos2021induction} in the \emph{DeliverCoffee} and \emph{PatrolAB} tasks as partial policies in the more complex \emph{DeliverCoffeeAndMail} and \emph{PatrolABC} tasks, respectively. In this way, we assess the sampling efficiency of our methodology when generalizing to longer planning horizons and more sub-goals. 
Figure \ref{fig:office} shows the performance of SR-DQN and the baselines. For both tasks, we tuned the base DQN algorithm to solve the easier setting (i.e. \emph{DeliverCoffee} and \emph{PatrolAB}) and then used the same set of hyperparameters to train all the agents in the more challenging tasks. On average, SR-DQN achieves the highest return by the end of training, also proving to be more stable with a lower standard deviation with respect to DQN in particular. On the other hand, RM-DQN converges more slowly to a lower average return with larger variance, proving the inefficiency of reward augmentation, as theoretically suggested by \cite{cheng2021heuristic}.
% and outperforming both standard DQN and RM-DQN, proving to be more robust in the presence of inaccurate or incomplete policies.

\begin{table}[ht]
\centering
\small
\caption{Execution times of DQN and SR-DQN algorithms on DoorKey (DK) and OfficeWorld (OW) tasks. Last column shows the time increment introduced by symbolic reasoning.}
\label{tab:execution_times}
\begin{tabular}{lcccc}
\toprule
 \textbf{Domain} & \textbf{Steps} & \textbf{DQN} & \textbf{SR-DQN} & \textbf{Increment} \\
\midrule
DK 8x8, 1 Key & 3M & 1h20mins & 1h24mins & 4 mins \textbf{(5\%)} \\
DK 8x8, 2 Keys & 5M & 2h10mins & 2h15mins & 5 mins \textbf{(3.85\%)} \\
DK 8x8, 4 Keys & 5M & 2h10mins & 2h15mins & 5 mins \textbf{(3.85\%)} \\
DK 16x16, 1 Key & 10M & 2h46mins & 2h54mins & 8 mins \textbf{(4.82\%)} \\
DK 16x16, 2 Keys & 10M & 2h46mins & 2h54mins & 8 mins \textbf{(4.82\%)} \\
OW, Deliver & 250k & 38mins30s & 39mins & 30s \textbf{(1.3\%)} \\
OW, PatrolABC & 250k & 40mins & 41mins & 1min \textbf{(2.5\%)}\\
\bottomrule
\end{tabular}
\end{table}

\subsubsection{DoorKey}
For \emph{DoorKey}, we evaluate performance across a range of scenarios with increasing complexity. We begin by scaling up the environment from a $5 \times 5$ to an $8 \times 8$ grid, and simultaneously increase the number of keys (with distinct colors) present in the map, either 2 or 4. We further extend the evaluation to a larger $16 \times 16$ grid, considering both one-key and two-key configurations\footnote{We omit the 4-keys configuration in the 16x16 map, as all tested algorithms, including ours, exhibit similarly low performance in this setting.}. In these settings, the agent must also correctly decide which key to pick, depending on the door's color, in order to finish the task soon and maximize the return.
These scenarios provide a compelling demonstration of the benefits of our neuro-symbolic approach, which effectively leverages prior knowledge to generalize to more complex domains. 
Importantly, none of these larger map configurations were considered in \cite{hazra2023deep}, the source of the $\pi_{ASP}$ policy.

% \begin{figure}
%     \centering
%     \includegraphics[width=0.9\linewidth]{images/dk8x8.png}
%     \caption{\emph{DoorKey} results on $8 \times 8$ grid with one door and key.}
%     \label{fig:doorkey8x8} 
% \end{figure}

\begin{figure*}
    \includegraphics[width=0.32\linewidth]{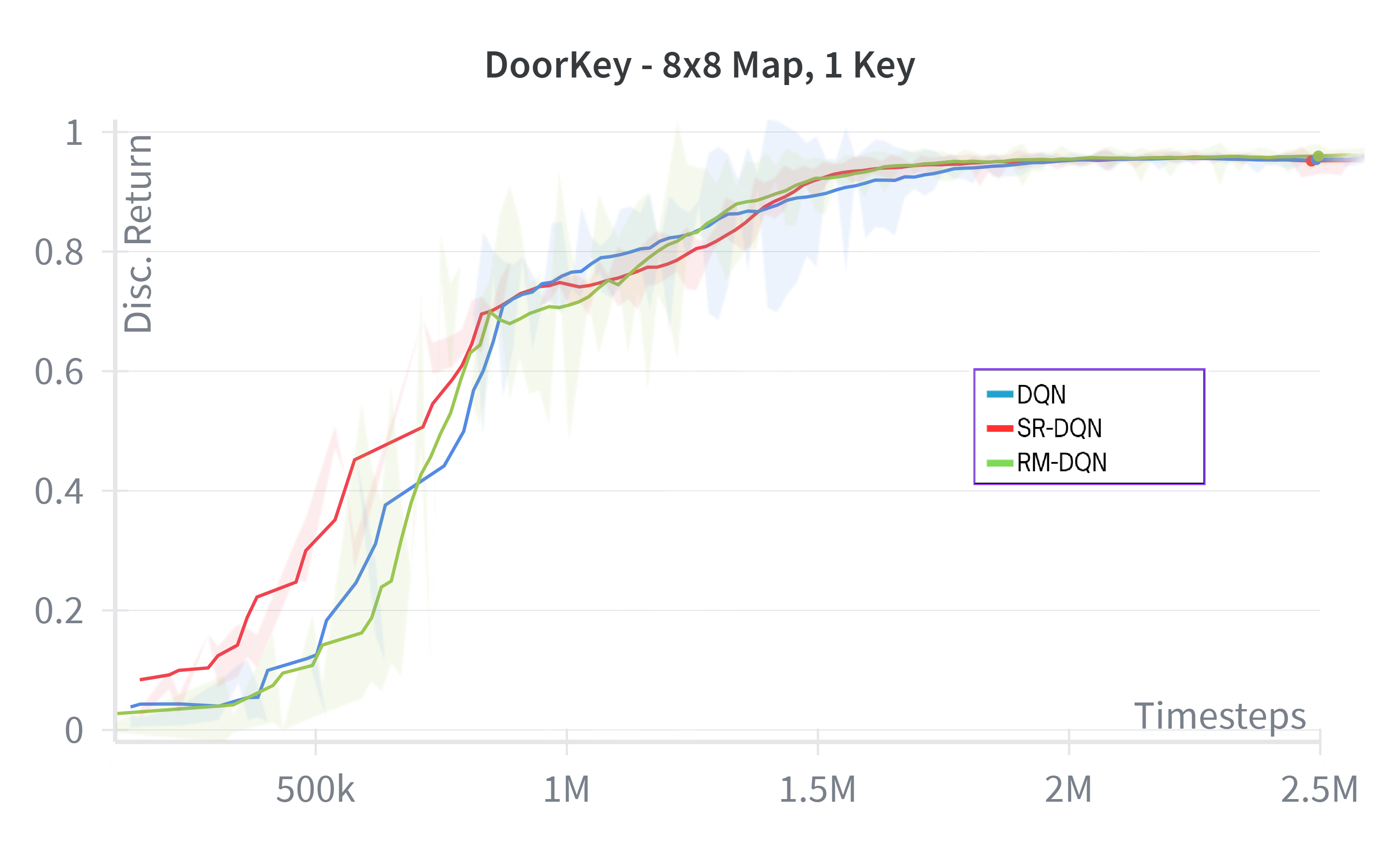}
    \includegraphics[width=0.32\linewidth]{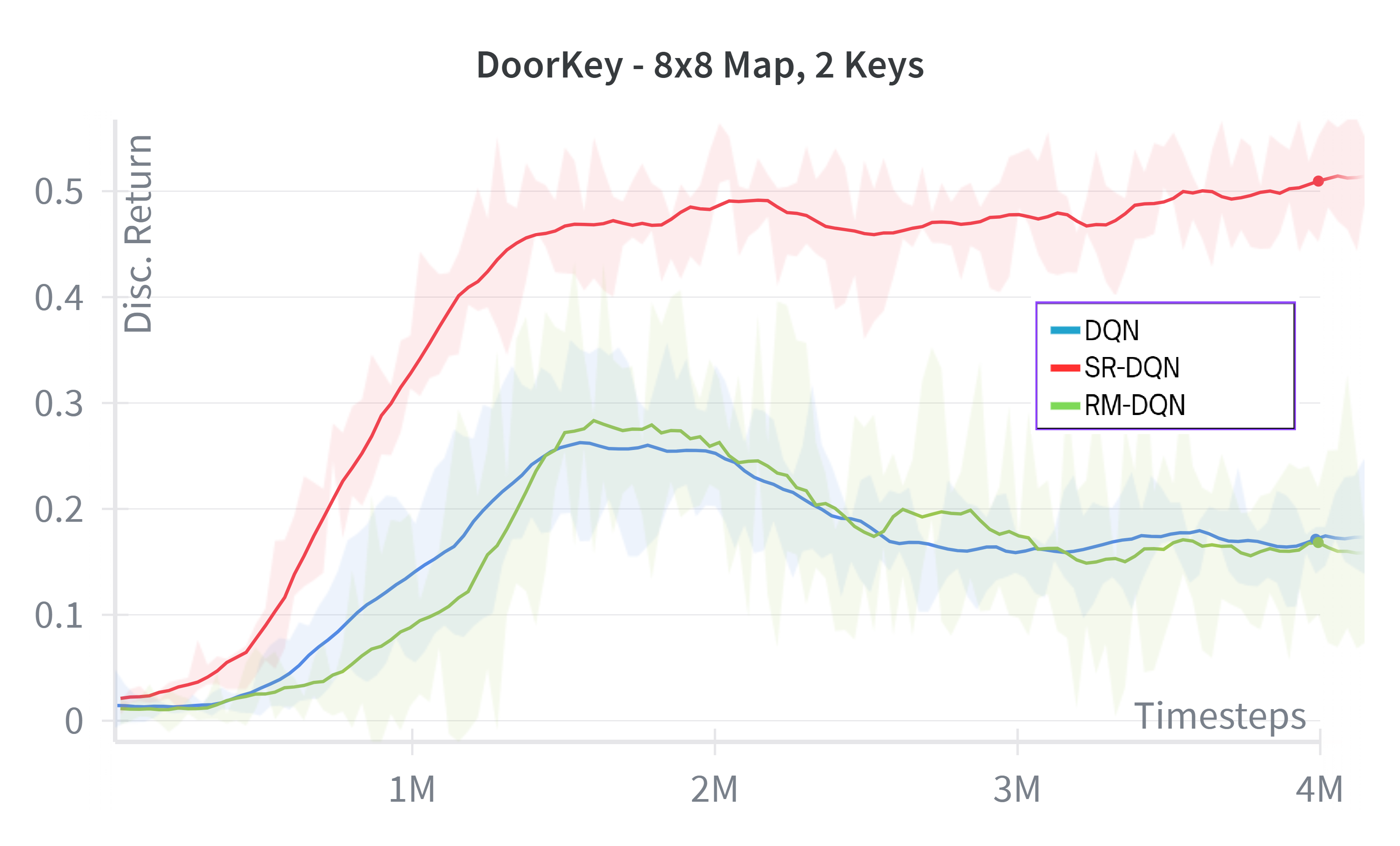}
    \includegraphics[width=0.32\linewidth]{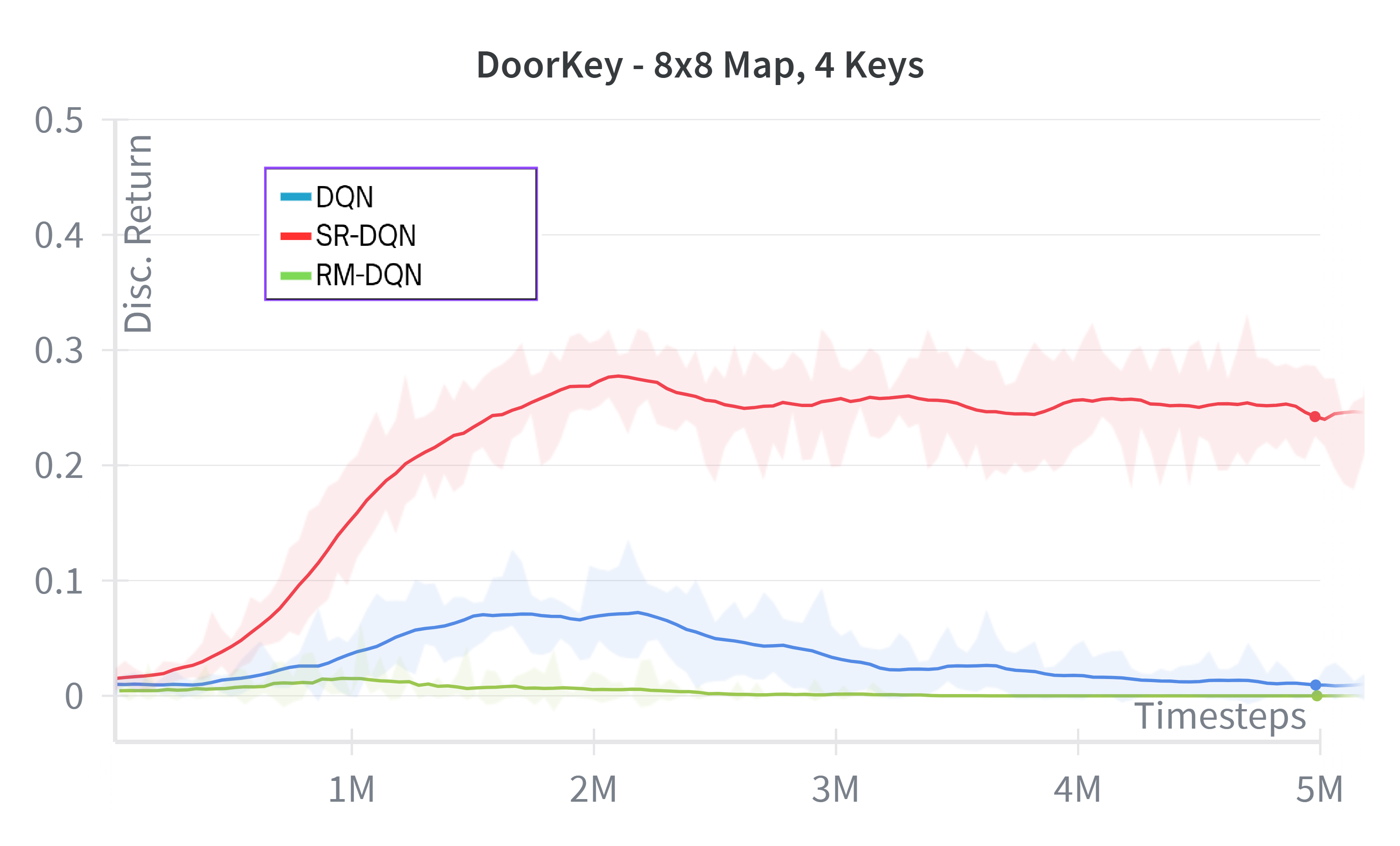}
    \\
    \vspace{1em}
    
    \includegraphics[width=0.32\linewidth]{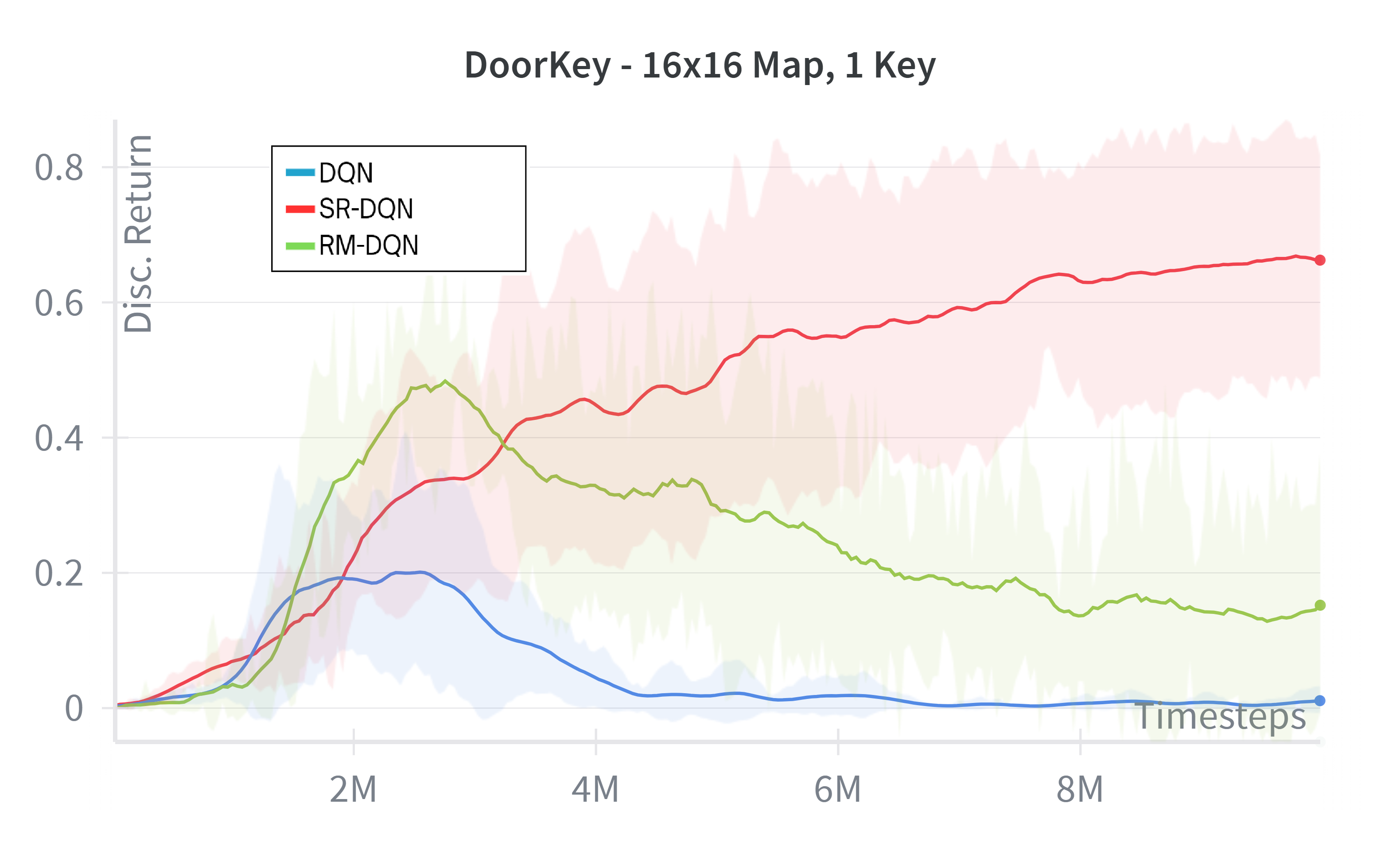}
    \hspace{1cm}
    \includegraphics[width=0.32\linewidth]{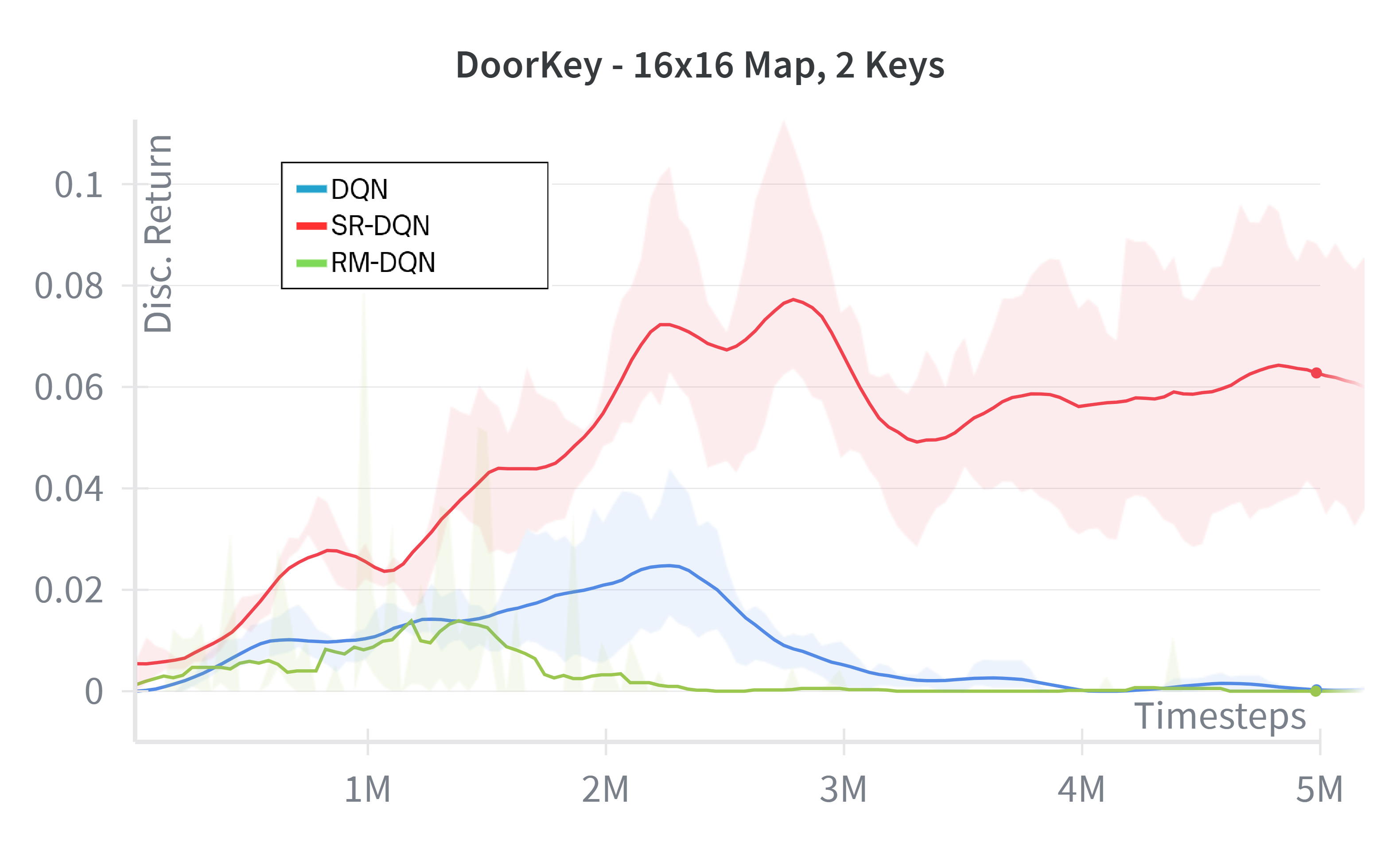}
    \caption{Training results on the DoorKey environment in random maps, varying grid size and number of keys.}
    \label{fig:doorkey8x8}
\end{figure*}

Figure \ref{fig:doorkey8x8} shows the performance of our algorithm and the baselines in a $8\times 8$ map with different amounts of keys in the environment. Even though all algorithms perform similarly in the easiest configuration with just one key (Figure \ref{fig:doorkey8x8}, top-left), the SR-DQN algorithm clearly outperforms both the baselines in the more challenging tasks, in which either 2 (Figure \ref{fig:doorkey8x8}, top-center) or 4 (Figure \ref{fig:doorkey8x8}, top-right) keys are present in the map.  
Our SR-DQN performs significantly better than both DQN and RM-DQN, showing a higher average return (more than two times the one obtained by DQN). 
Finally, Figure \ref{fig:doorkey8x8} (bottom line) clearly shows that, even in the bigger $16\times16$ maps, our algorithm outperforms both the baselines, being the only one able to obtain an acceptable return in both scenarios.

Overall, our neurosymbolic integration demonstrates clear improvements over RM-DQN in environments with longer planning horizons (e.g., multiple keys or larger grids), highlighting the limitations of reward augmentation or shaping in such scenarios.

Finally, in Table \ref{tab:execution_times}, we report the execution times of DQN and SR-DQN across all tested tasks. It is evident that the additional overhead introduced by the symbolic inference, indicated in the last row, has a negligible impact on the overall training duration. This demonstrates that integrating symbolic guidance does not compromise the efficiency of the learning process while still providing the benefits evidenced throughout this section.

%  \begin{figure}
%     \centering
%     \includegraphics[width=0.9\linewidth]{images/dk8x84k.png}
%     \caption{Results on the 8x8 Doorkey environment with 4 keys}
%     \label{fig:doorkey4keys}
% \end{figure}

% \begin{figure*}[ht]
%     \centering
%     \includegraphics[width=0.36\linewidth]{images/dk16x161k.png}
%     \hspace{1cm}
%     \includegraphics[width=0.36\linewidth]{images/dk16x162k.png}
%     \caption{Training results on the Doorkey environment in random 16x16 maps with 1 (left) and 2 (right) keys.}
%     \label{fig:doorkey16x16}
% \end{figure*}

\subsection{Ablation study}

\begin{figure*}[ht]
    \centering
    \includegraphics[width=0.32\linewidth]{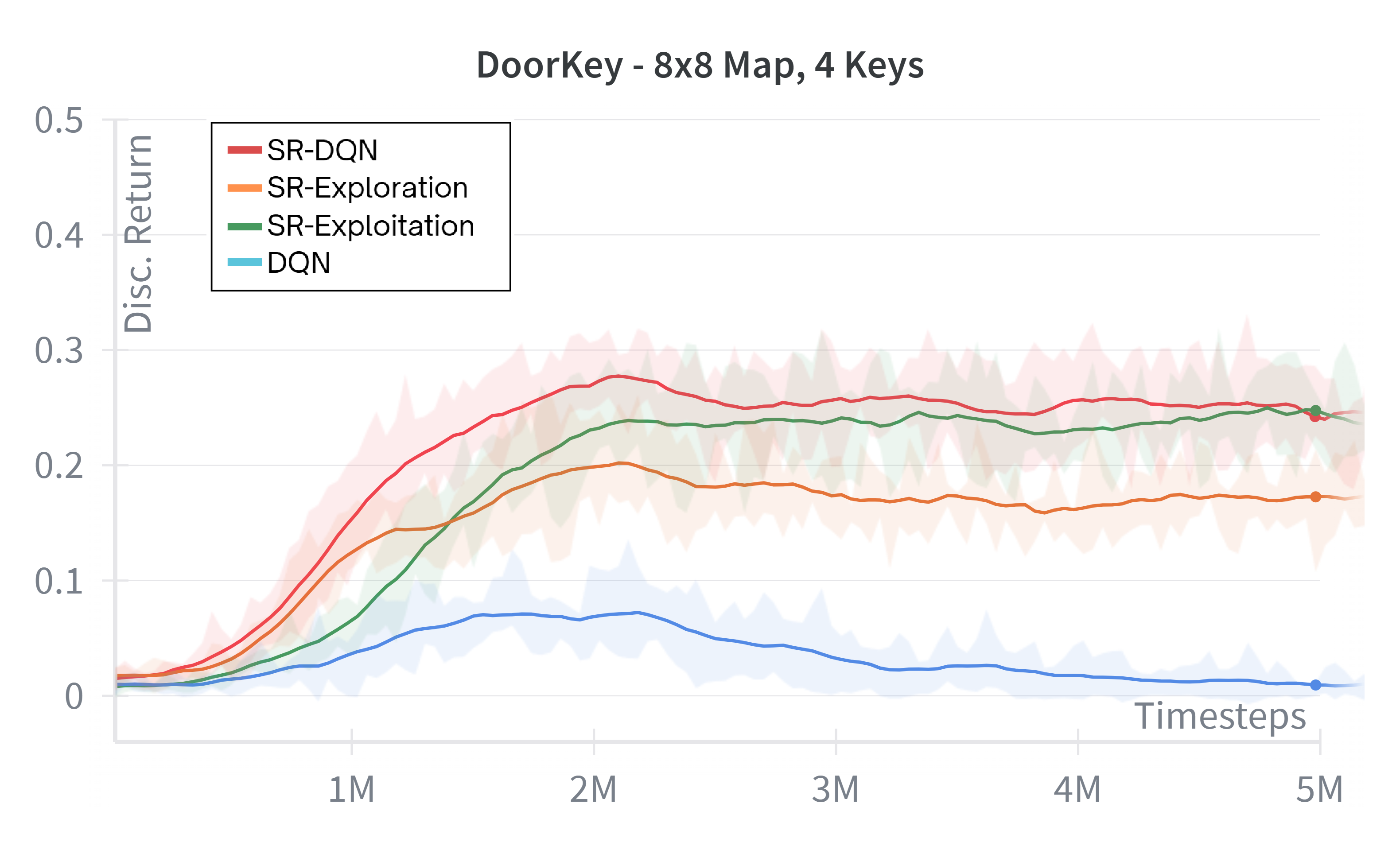}
    \includegraphics[width=0.32\linewidth]{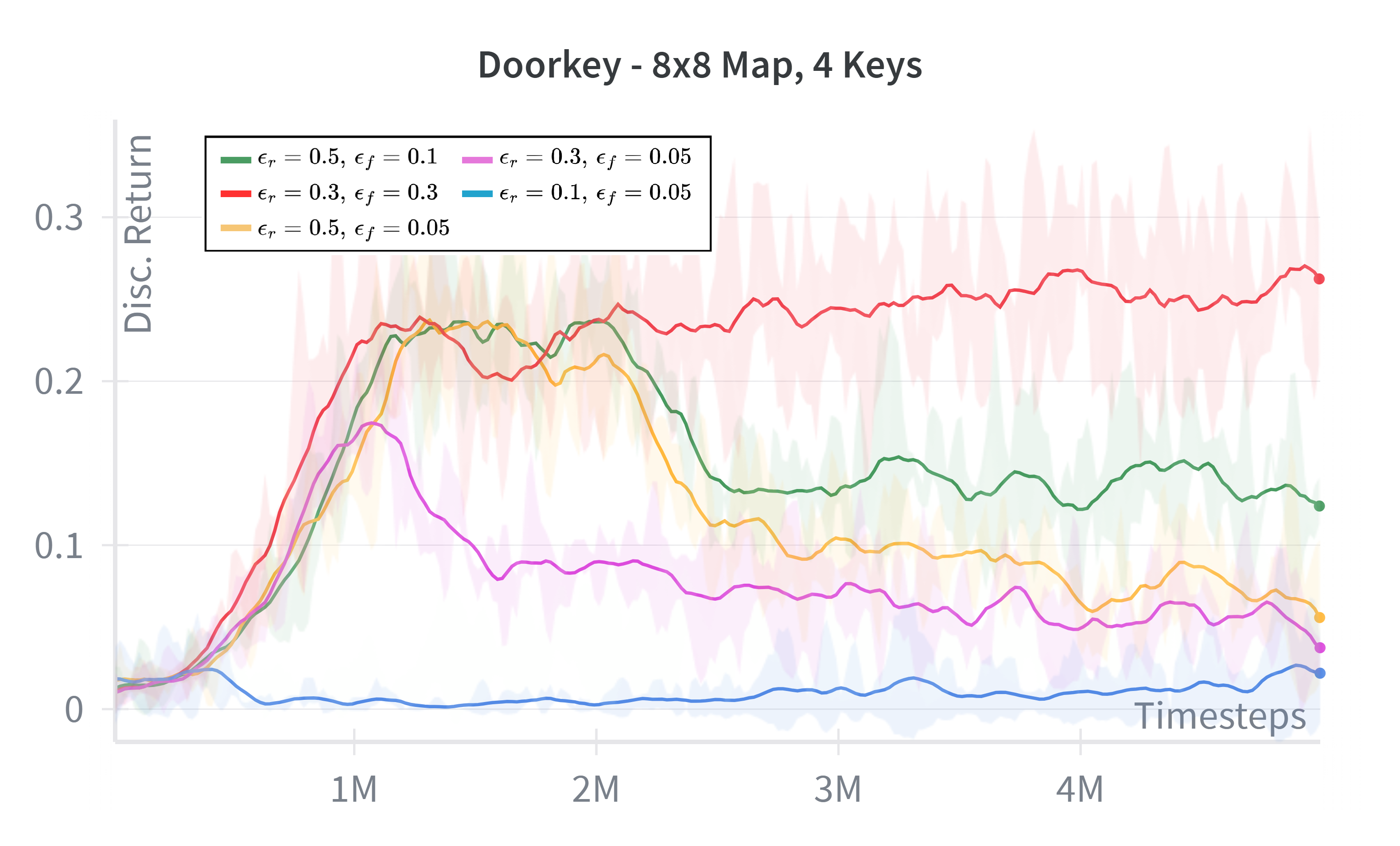}
    \includegraphics[width=0.32\linewidth]{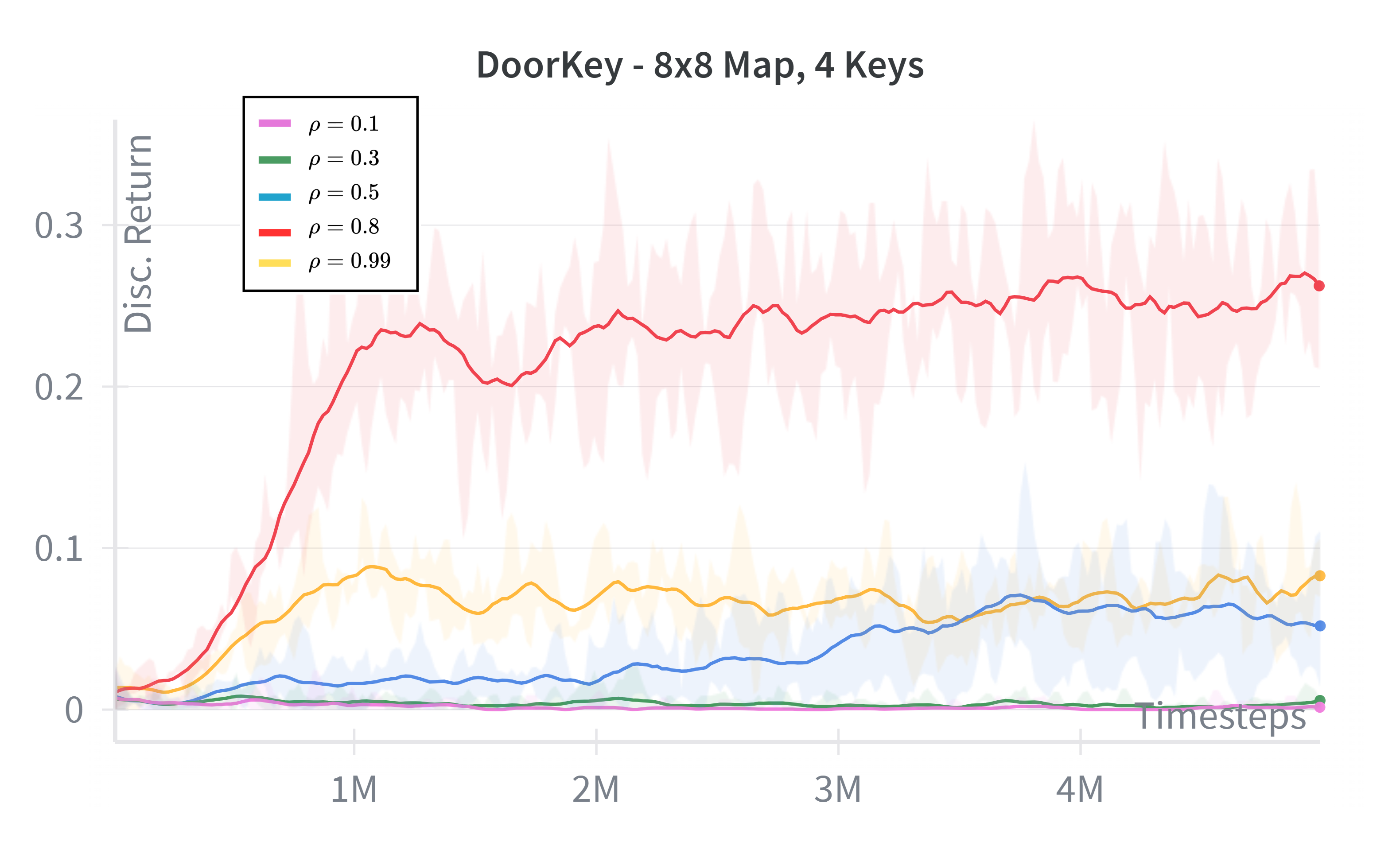}
    
    \caption{Ablation study over the different components of the SR-DQN algorithm, namely SR-Exploration and SR-Exploitation, compared to the baselines and the full SR-DQN algorithm (left) and training curve of SR-DQN algorithm with either different $\epsilon_f$ and $\epsilon_r$ (center), or different $\rho$ values (right). All studies are performed on $8\times8$ DoorKey maps with 4 keys.}
    \label{fig:ablation}
\end{figure*}

Our SR-DQN (Algorithm \ref{alg:sr_dqn}) combines symbolic reasoning both in the exploration (Algorithm \ref{alg:sr_exp}) and the exploitation (Algorithm \ref{alg:sr_qvals}) phases of DRL, modulated by the decay law of $\epsilon$ (Line 20). Together with $\rho$, the values of $\epsilon_r$ and $\epsilon_f$ determine the impact of $\pi_{ASP}$ on the training loop.
We now want to investigate in more detail the role of these components independently.

Figure \ref{fig:ablation} (left) shows the performance of both SR-Exploration (Algorithm \ref{alg:sr_exp}) and SR-Exploitation (Algorithm \ref{alg:sr_qvals}) when employed as the only symbolic component within the full Algorithm \ref{alg:sr_dqn}. Specifically, as in standard DQN, when disabling SR-Exploration, we sample $a$ uniformly from $A$ during exploration; for SR-Exploitation, we do not rescale Q-values at Line 3 of Algorithm \ref{alg:sr_qvals}.
We perform this ablation study on the same $8\times8$ Doorkey environment with 4 keys, which provides a challenging yet tractable scenario where good performance can still be achieved. Both SR-Exploration and SR-Exploitation alone outperform standard DQN, but SR-Exploitation alone achieves performance very close to that of the full SR-DQN, underscoring the importance of value shaping as a key contributor to overall performance. In contrast, SR-Exploration alone provides a smaller improvement but a faster growth of the return in the very beginning of the training. 

In Figure \ref{fig:ablation} (center), we instead keep the full SR-DQN algorithm, but vary the values of $\epsilon_f$ and $\epsilon_r$. Starting from an equal initial value $\epsilon_i = 1$, in this way we modify the decremental behaviour of $\epsilon$, thus varying the impact of the symbolic component of our algorithm.
We perform this test in the \emph{DoorKey} environment with $8 \times 8$ grid and 4 keys, which represents a great challenge for all the tested baselines but still gives the chance to learn good policies.
The optimal curve corresponding to Figure \ref{fig:doorkey8x8} (right) is reported in red ($\epsilon_f=0.3, \epsilon_r=0.3$).
Figure \ref{fig:ablation} (center) shows that, when decreasing $\epsilon_f$ and $\epsilon_r$, thus reducing the impact of the symbolic policy, the training performance decreases significantly. Moreover, a too high $\epsilon_f$ value results in unstable policies, thus gaining worse returns.

Finally, Figure~\ref{fig:ablation} (right) shows the performance of SR-DQN under different values of the confidence parameter~$\rho$. 
Both excessively low ($\rho \leq 0.5$) and excessively high ($\rho$ close to~1) confidence values lead to suboptimal performance. 
In the former case, the symbolic component exerts too little influence, preventing the agent from effectively exploiting the structured prior knowledge. 
In the latter, too much reliance is placed on the symbolic rules, whose accuracy is limited since they are learned from simplified versions of the tasks. 
This analysis highlights the need for a balanced integration between neural and symbolic components, where~$\rho$, $\epsilon_r$, and $\epsilon_f$ regulate the trust in imperfect symbolic knowledge.

\section{Conclusion and Future Work}
We presented SR-DQN, a novel neuro-symbolic DRL approach to tackle the problems of scalability and sampling inefficiency in DRL, in environments with long planning horizons, sparse rewards, and multiple sub-goals. 
Our methodology exploits partial logical policy specifications representing the optimal strategy in easy-to-solve domain instances with limited planning horizon. Then, we perform automated reasoning to entail suggested actions from the logical specifications, biasing both the exploration phase of $\epsilon$-greedy DRL agents and the Q-values produced by the neural component during training, to encourage the choice of promising symbolic actions. We exploit an $\epsilon$-decay schedule to balance symbolic reasoning and neural learning over time. Importantly, the added symbolic component doesn't represent a significant computational overhead for the original DRL algorithm.
We empirically demonstrated the benefits of SR-DQN in two benchmarks, \emph{OfficeWorld} and \emph{DoorKey}, both of which present the challenges mentioned above, as well as partial observability in larger maps. SR-DQN consistently outperformed all the selected baselines (namely, standard DQN, DQN with reward machines, which represent a state-of-the-art technique in neuro-symbolic DRL), also being the only method capable of achieving significant returns in challenging, partially observable \emph{DoorKey} tasks with more items (e.g., multiple keys) and sub-goals, where all tested baselines performed much worse. 
%An ablation study evidences the synergistic importance of both symbolic exploration and exploration on the final performance.

In future work, we plan to generalize our methodology to a broader class of DRL algorithms beyond $\epsilon$-greedy strategies. This includes integrating our framework with policy gradient and actor-critic methods. Additionally, we aim to extend our approach to more expressive logical representations, such as temporal or probabilistic logic.

%%%%%%%%%%%%%%%%%%%%%%%%%%%%%%%%%%%%%%%%%%%%%%%%%%%%%%%%%%%%%%%%%%%%%%%%

%%% The acknowledgments section is defined using the "acks" environment
%%% (rather than an unnumbered section). The use of this environment 
%%% ensures the proper identification of the section in the article 
%%% metadata as well as the consistent spelling of the heading.

%%%%%%%%%%%%%%%%%%%%%%%%%%%%%%%%%%%%%%%%%%%%%%%%%%%%%%%%%%%%%%%%%%%%%%%%

%%% The next two lines define, first, the bibliography style to be 
%%% applied, and, second, the bibliography file to be used.

\bibliographystyle{ACM-Reference-Format} 
\bibliography{sample}

%%%%%%%%%%%%%%%%%%%%%%%%%%%%%%%%%%%%%%%%%%%%%%%%%%%%%%%%%%%%%%%%%%%%%%%%

\end{document}